# A new database of Houma Alliance Book ancient handwritten characters and classifier fusion approach

Xiaoyu Yuan[1], Zhibo Zhang[1], Yabo Sun[1], Zekai Xue[1], Xiuyan Shao[2], Xiaohua Huang[1,3]*
1. School of Computer Engineering, Nanjing Institute of Technology
2. Southeast University, Nanjing
3. Jiangsu Province Engineering Research Center
of IntelliSense Technology and System, Nanjing, Jiangsu
* Corresponding author

**ABSTRACT**

The Houma Alliance Book is one of the national treasures of the Museum in Shanxi Museum Town in China. It has great historical significance in researching ancient history. To date, the research on the Houma Alliance Book has been staying in the identification of paper documents, which is inefficient to identify and difficult to display, study and publicize. Therefore, the digitization of the recognized ancient characters of Houma League can effectively improve the efficiency of recognizing ancient characters and provide more reliable technical support and text data. This paper proposes a new database of Houma Alliance Book ancient handwritten characters and a multi-modal fusion method to recognize ancient handwritten characters. In the database, 297 classes and 3,547 samples of Houma Alliance ancient handwritten characters are collected from the original book collection and by human imitative writing. Furthermore, the decision-level classifier fusion strategy is applied to fuse three well-known deep neural network architectures for ancient handwritten character recognition. Experiments are performed on our new database. The experimental results first provide the baseline result of the new database to the research community and then demonstrate the efficiency of our proposed method.

## 1. INTRODUCTION

Handwritten Chinese character recognition has attracted extensive attention since the 1960s. Since then, some Chinese character databases and recognition have been developed for Chinese character recognition. For example, the CASIA-OLHWDB1.0 dataset established by the Chinese Academy of Sciences in 2011 contains 420 sets of handwriting samples. Each set contains 171 alphanumeric symbols, 3866 commonly used Chinese characters (3740 of which belong to GB2312-level Chinese characters), for a total of 1,694,741 valid samples [1]. With the development of deep learning, Chinese character recognition has rapidly developed and is widely used in optical character recognition [2-4], but it still does not investigate ancient Chinese handwriting. In the last decade, the Houma alliance book ancient handwriting [5], an ancient Chinese character, has been investigated in the historical cultural field. According to Chinese history, the Houma alliance script originated in the Spring and Autumn period and the Warring States period. It was written in the Jin State in the Eastern Zhou Dynasty to record the oath. To the best of our knowledge, this is the earliest and largest number of calligraphy documents to date. Compared with modern printed Chinese characters or our handwritten Chinese characters, the handwriting of the ancient book has been severely corroded in the last 2,000 years. On the other hand, the Houma Alliance book ancient handwriting character lacks standard writing rules and has the diversity of character shapes, which means one character may have different ways to write. In addition, the lack of a standard, large and free database is one of the main obstacles to the research in Houma Alliance book ancient handwriting recognition. The difficulty comes from the complexity of text segmentation and database annotation.

To date, there are few works on Houma alliance book ancient handwritten recognition. We briefly review the state-of-the-art works on handwritten Chinse character recognition. Due to the diversity of handwriting styles and many character classes, handwritten Chinese character recognition has still been investigated by researchers. Casey et al. first released the work about printed Chinese character recognition [6], in which they used a simple template matching method to recognize 1,000 printed Chinese characters. According to the type of data acquisition, handwritten Chinese character recognition can be divided into two categories: offline and online. Offline handwritten character recognition consists of the writer writing the words on paper in advance, which is then converted into a text image by a scanner and then recognized by the computer as the corresponding words. Normally, the offline procedure consists of image normalization, feature extraction and classifier training. With the development of deep learning technology, deep neural network architectures have been widely used in the handwritten character recognition. For example, a multi-column deep neural network was proposed to recognize Chinese characters [7]. Online handwritten character recognition is carried out by the machine according to the stroke, stroke order and other characteristics of the written words while the person is writing them. In [8], a stroke sequence-dependent deep convolutional neural network method was proposed for online handwritten Chinese characters, in which they exploited stroke sequence information and eight-directional features of Chinese characters. Deep neural networks have been successfully implemented in this domain. On the other hand, in various application areas of pattern recognition, combining multiple classifiers is regarded as a method for achieving a substantial gain in performance of systems. In [9] Lin et al. combined multiple classifiers based on statistics for handwritten Chinese character recognition. Motivated by their work, we propose fusing multiple deep neural networks to recognize Houma Alliance Book ancient handwritten characters.

In summary, in this paper, we focus on establishing a new database of Houma Alliance Book ancient handwriting and design a new system to recognize them. To the best of our knowledge, this is the first study to investigate Houma Alliance Book ancient handwriting in the computer vision community. This paper makes two contributions. First, we collected the recognized Houma alliance letter glyphs according to the report documents of archaeologists, and based on this, we expanded the sample data by human imitative writing. Second, we present an algorithm based on a decision-level fusion strategy to identify ancient characters.

The rest of this paper is organized as follows: Section 2 introduces the literature review of the Houma Alliance book and the ancient character database. In Section 3, we describe the process of establishing the Houma Alliance handwriting database. In Section 4, we introduce our proposed method for Houma alliance handwriting recognition. In Section 5, we present experiments on the new database and discuss the results of different models. Finally, a conclusion is given in Section 6.

## 2. RELATED WORK

### 2.1 Houma Alliance Book

The Houma Alliance Book was discovered in 1965 by the Shanxi Provincial Cultural Heritage Commission during excavations at the Jin site of Houma, Shanxi, and excavated from November to May of the same year. This unprecedented discovery not only advanced the development of Chinese writing but also filled many gaps in written sources. The Houma Alliance Book was written on jade tablets with a brush, and the handwriting is generally vermilion, with a few in black. The script is similar to that of late Spring and Autumn Period time bronze inscriptions. The special feature of the Houma alliance book lies in its complete and systematic content of the alliance rhetoric. Wang et al. collated and studied this collection of alliance books [10]; in the early 1970s, the Shanxi Provincial Committee of Cultural Relics also copied, collated and studied this collection of materials and published the Houma Alliance Book [11].

As a living specimen of ancient Chinese calligraphy, the Houma Alliance Book records the history of numerous Spring and Autumn Period times, and it becomes particularly important to recognize and understand the contents of the Houma Alliance Book. With the continuous development and optimization of digital image processing technology in the restoration of cultural relics, Lin et al. proposed using suprathreshold stochastic resonance for the virtual restoration of Houma convenant tablets [12]. Thanks to restoring inscriptions, interpreting the Houma Alliance Book can be analyzed through inscription text restoration and inscription color restoration.

### 2.2 Ancient character handwriting databases

Large databases have been developed for the recognition of handwriting based on ancient scripts, such as Arabic historical subwords and oracle bone inscriptions, which have helped researchers develop artificial intelligence-based algorithms to assist in the research of ancient scripts. For example, in [13], Zoizou *et al.* present a MOJ-DB database of Arabic historical subwords, which contains 560,000 subwords distributed on 5,600 different classes. It was built using 64 pages extracted from 10 books written in the $17^{th}$ and $16^{th}$ centuries. In [14], Li *et al.* established a handwriting oracle bone character containing 83,245 character-level samples that are grouped into 3,881-character categories. In [15], Narang *et al.* presented optical character recognition based on SIFT and Gabor for the recognition of the Devanagari ancient manuscript. A database was collected, including 5,484 samples of Devanagari characters from various ancient manuscripts placed in libraries and museums. In [16], Zhao *et al.* used a multilayer adversarial neural network with a Laplacian structure in character recognition technology for Shui characters in ancient books. In [17], Xu *et al.* built a database by annotating 11,937 pages of ancient Chinese handwritten documents, which belong to 10,350 categories.

In contrast, the ancient script of the Houma alliance book has been so forgotten that there is no publicly available database in recent years. The archaeological report on the Houma Alliance Book [18], compiled by the Shanxi Provincial Cultural Relics Working Committee, gave this study a sample of over 5,000 related jade tablets obtained from archaeology. The Houma Alliance Book - Character List (Revised) written by Zhang *et al.* [11] and its copy provide a paper version of the text template. Their research is dedicated to the initial work of collating and building a library of the character forms of the League Book, which will provide a good foundation for further research on the Houma League Book.

## 3. HOUMA ALLIANCE BOOK ANCIENT HANDWRITTEN CHARACTER DATABASE

### 3.1 Acquisition details

We acquired identified ancient handwritten characters and corresponding traditional Chinese characters from the archaeological report of the Houma Alliance Book [18]. According to the archaeological report of the Houma Alliance Book, we initially collect and collate a small dataset of 297 identified characters, each of which has multiple variants, including 97 types of the Chinese character "敢". A total of 1,665 original images are collected from the archaeological report of the Houma Alliance Book, with the existing copies of the archaeological report cropped to a size of $64 \times 64$ pixels. To augment the original database, based on the iPad digital handwritten tool, four persons are asked to imitate the Houma Alliance Book ancient characters. Lastly, this database totally contains 3,547 images and 297 classes. Its data distribution is shown in Figure 1. It is found that there was a wide disparity in the number of characters in each of the 297 categories, with almost 90.5% of the characters having no more than 20 characters, and 188 of them having less than 5 characters; a very small number of characters had more than 40 variants.

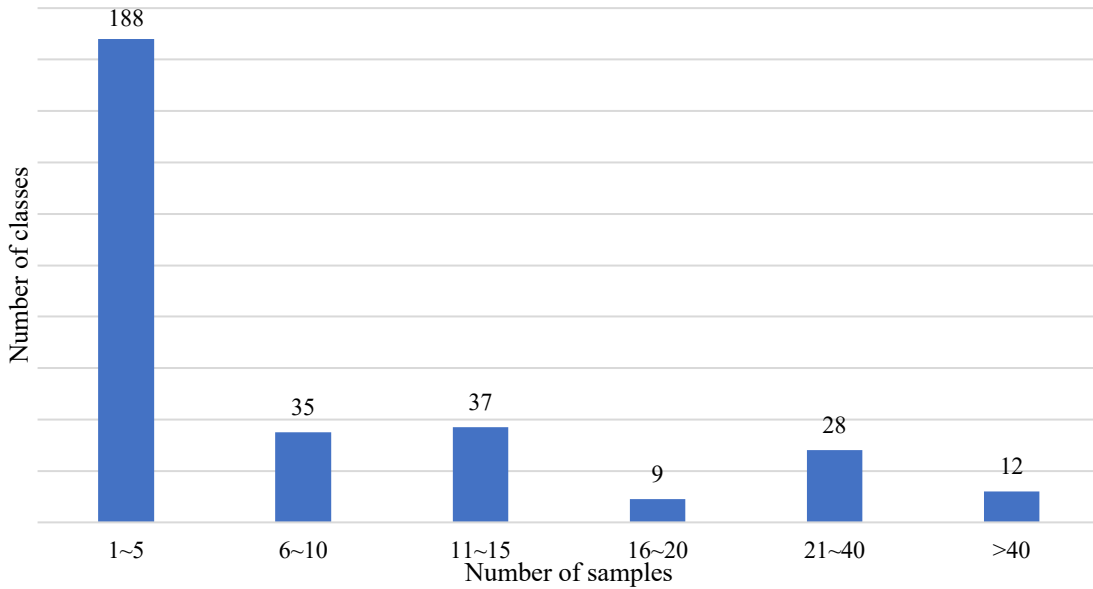

Figure 1. Statistical analysis of the data distribution in terms of the number of samples, where the x-axis represents the range of the number of samples, and the y-axis is the number of classes.

## 3.2 Data annotation

As the data collection relies on the 297 identified ancient character classes published in the archaeological report of the Houma Alliance Book and as the identification results in the archaeological report of the Houma Alliance Book correspond mostly to traditional Chinese characters and facing the problem that some of the less common characters are now non-existent, this paper uses a special character compilation program shown in Figure 2(a) to create data labels by means of character building. Based on annotation, we construct 3,574 images and their annotation in the form of traditional Chinese characters. Some examples are shown in Figure 2(b).

To easily train the system, we use a trick to label samples. In this paper, we use a standard index format of naming $P\_I\_S\_x$, where $P$ is the number of pages in its monogram table, $I$ is the position serial number of the word, $S$ is the sample serial number of its typeface, and $x$ is the serial number of that handwritten sample. Based on the new annotation format, some examples are shown in Figure 3. Therefore, our database has two annotations. The annotation details are reported in Table 1. It describes the ancient Chinese characters of the Houma Alliance Book based on the 297 classes of ancient Chinese characters collected, and the annotation table should have fields for index and words.

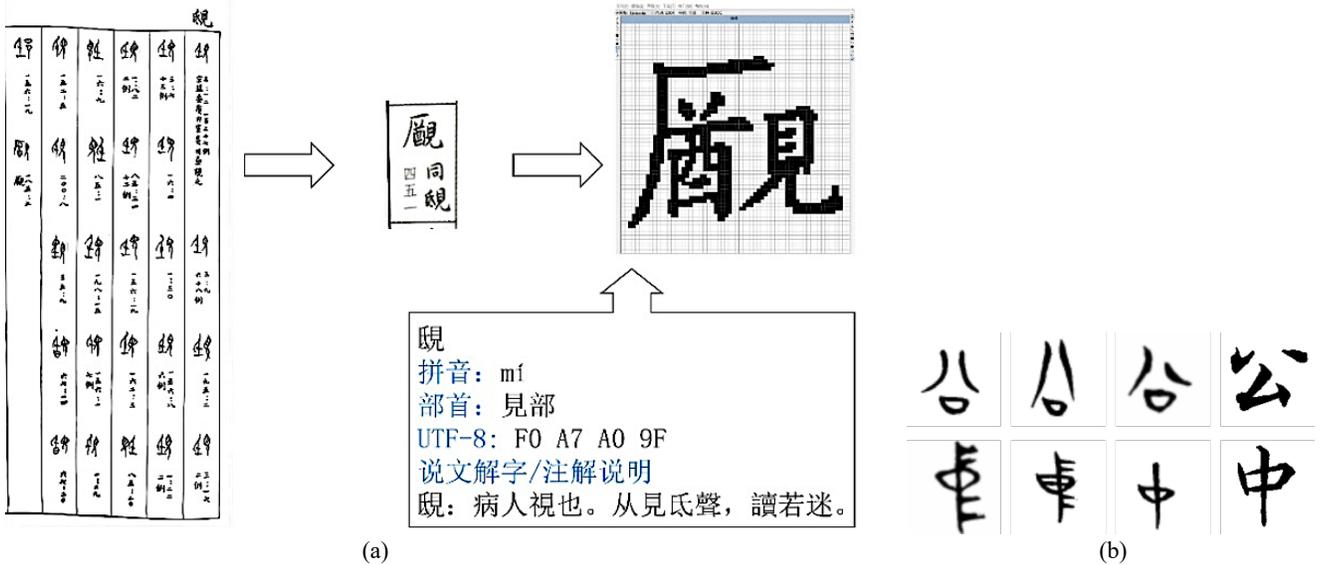

(a)         (b)

Figure 2. Data annotation procedure. (a) One example of annotating Houma Alliance Book ancient character. (b) Two examples acquired from [18], where the first three columns are Houma Alliance Book ancient characters, and the last one is their corresponding annotated traditional Chinese character.

Table 1. The annotation information includes the index and words.

| Annotation | Description |
|---|---|
| **Index** | $P\_I\_S\_x$, $P$ is the number of pages in its monogram table, $I$ is the position serial number of the word, $S$ is the sample serial number of its typeface, and $x$ is the serial number of that handwritten sample. |
| **Character** | The traditional Chinese character |

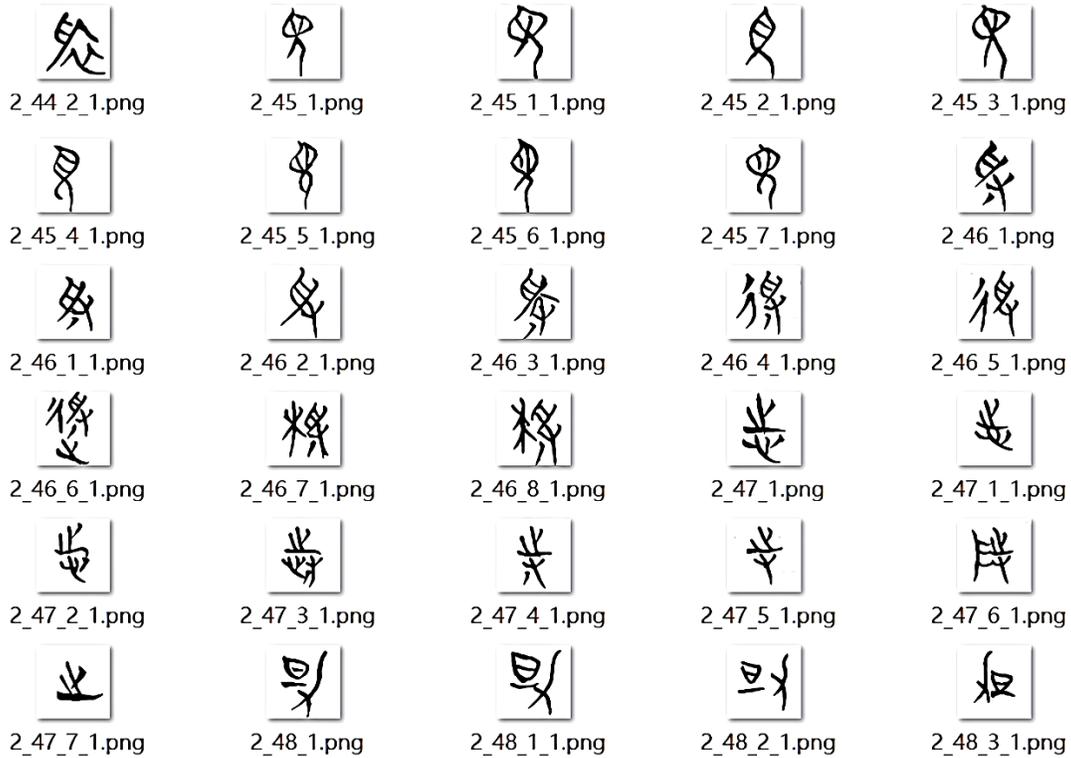

Figure 3. Annotation in terms of the index of Houma Alliance Book ancient handwritten characters.

## 4. METHODOLOGY

In this section we describe the proposed method for our newly collected database. In the following subsection, we briefly describe decision-level classifier fusion and three state-of-the-art deep architectures used in the fusion method for Houma Alliance Book ancient handwritten character recognition.

### 4.1 Decision-level classifier fusion

Classifier fusion has been presented on several types of classifiers to enhance the performance of individual classifiers. The previously proposed works [19, 20] demonstrated more precise, accurate, and certain aspects of any decision that constructs the fusion. Normally, with this context, a set of decisions is first conducted and then synthesized via a particular classifier fusion strategy.

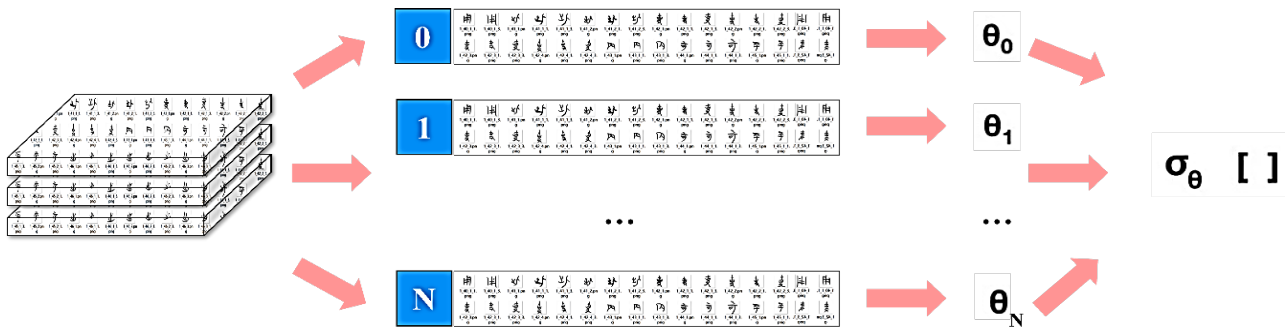

Figure 4. The architecture of Decision-level classifier fusion, where $N$ is the number of base architectures.

Our proposed algorithm is shown in Figure 4. Naïve Bayes (**NB**) combination is used to integrate the classifier label outputs. Mathematically, given a test image $x$, with $N$ base classifiers, the probability for the $l_n$ class predicted by the $n$th classifier is denoted as $\theta_{n,l_n}(x)$, where $l_n \in \{1, ..., C\}$, the conditional independence can be defined by

$$\theta_{n,l_n}(x|w) = \theta(l_1, ..., l_N|x_t, w) = \prod_{n=1}^{N} \theta_{n,l_n}(l_n|x, w), \quad (1)$$

where $w \in \{1, ..., C\}$ is the class label predicted by the classifier combination. The posterior probability required for the $n$th classifier can be explained by

$$\theta_{n,l_n}(w|x) = \frac{\theta(x)\theta(x|w)}{\theta(w)} = \frac{\theta(x)\prod_{n=1}^{N}\theta_{n,l_n}(l_n|x,w)}{\theta(w)}. \quad (2)$$

It is noted that $\theta(w)$ does not reply on $l_n$. Therefore, the final probability of $x$ is formulated as,

$$\sigma_\theta(x) = \max_{l_n} \theta(w_k) \prod_{n=1}^{N} \theta_{n,l_n}(w|x), \quad (3)$$

where the $max$ function is employed to choose the final class w.r.t. the maximum probabilities of all classes.

In our system, the base classifiers are LeNet [11], AlexNet [12], and ResNet [9]. The voting method obtains the result of the fusion model by training multiple classification models at the same time and merging and averaging. The voting method includes hard voting and soft voting. Hard voting obtains the fusion result by voting on the prediction results of various models while soft voting uses the possibility predicted by various models as weights to perform weighted voting based on hard voting. Generally, soft voting obtains better results than hard voting. Therefore, the soft voting method is used to perform decision-level classifier fusion (DCF) across LeNet, AlexNet, and ResNet. Finally, we obtain four ensembled models, denoted as LeNet+Alexnet (**DCF-LA**), LeNet+ResNet (**DCF-LR**), AlexNet+ResNet (**DCF-AR**), and LeNet+AlexNet+ResNet (**DCF-LAR**).

### 4.2 Base Classifier

This part describes the architecture of the three base classifiers.

LeNet, proposed by Yann et al. [21], consists of one input layer, three convolutional layers, two pooling layers, and two fully connected layers. The image of size $112 \times 112$ is fed into the input layer. There are three convolutional layers based on $5 \times 5$ filters, followed by mean pooling with $2 \times 2$ patches. The ReLU function is used for activation, as it leads to faster training. The Dropout Layer is added with a factor of 0.2 to deal with overfitting. The number of neurons of the last Dense layer is assigned to the number of classes, and the Softmax activation function obtains probabilities between 0.0 and 1.0. A pretrained LeNet model on ImageNet and the Adam learning optimization algorithm are implemented.

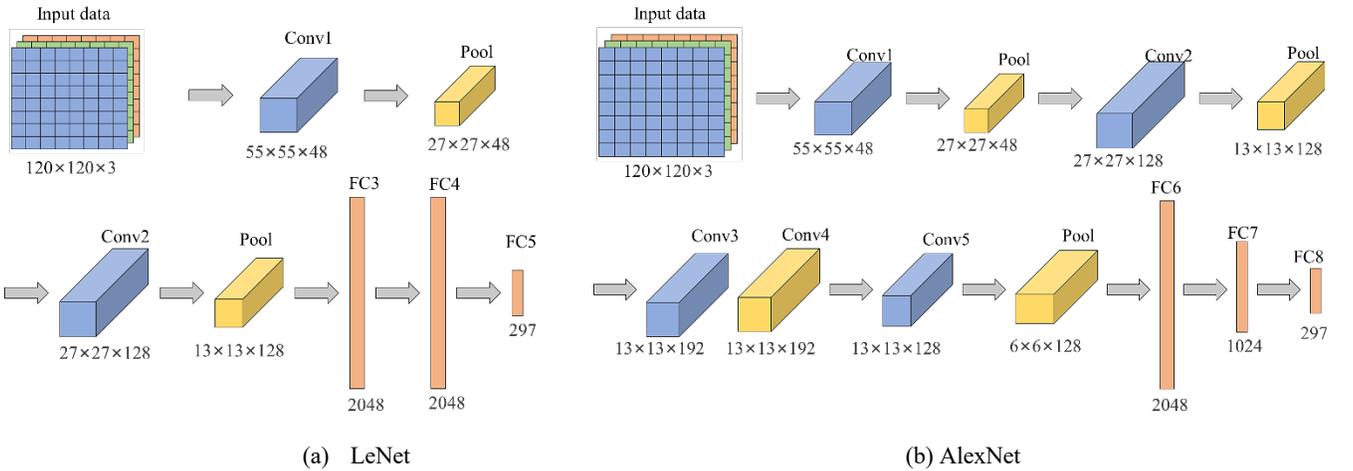

Figure 5. The framework of (a) LeNet and (b) AlexNet

AlexNet comprises of six convolution layers and three fully connected layers with a Softmax layer as the final layer. Those same convolution layers have normalization and pooling layers immediately after them, and traditionally, all the layers are initiated or activated using the rectifying linear unit (ReLU). We employ AlexNet pretrained on ImageNet, where the output of SoftMax is equal to the number of classes (297), instead of 1000-way. To reduce the overfitting in the fully connected layers, a Dropout layer with a factor of 0.2 is added. The Adam learning algorithm is employed to optimize AlexNet.

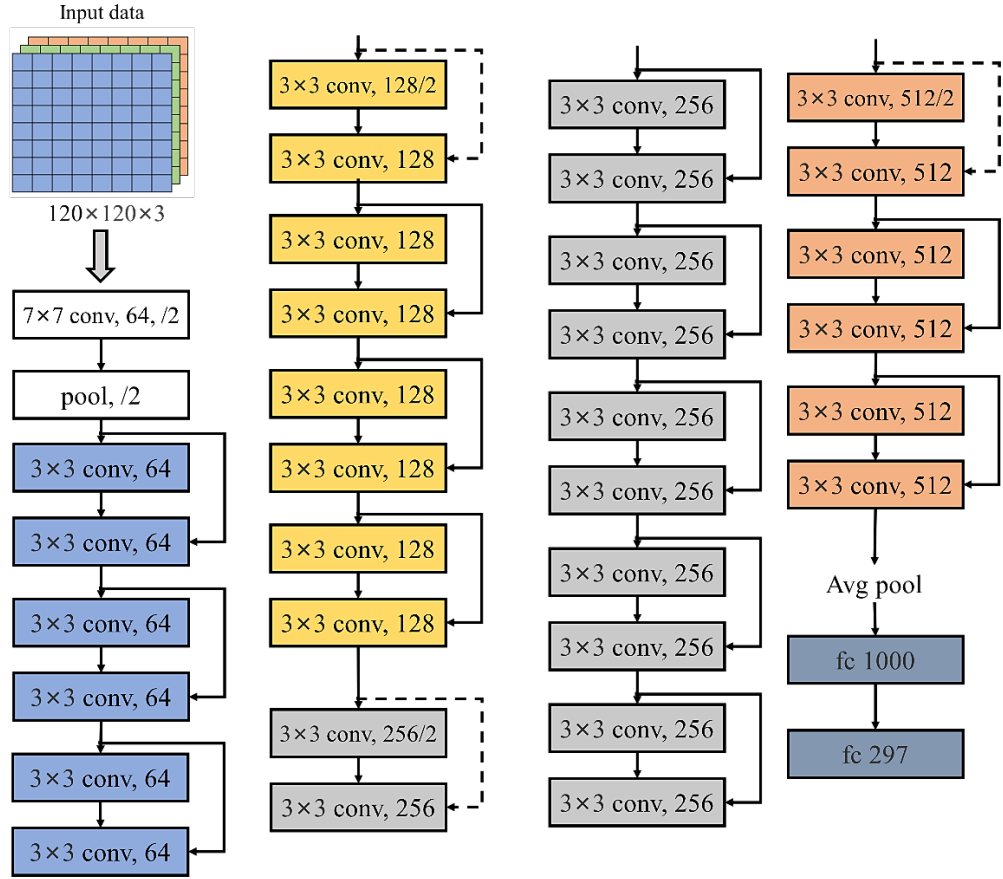

Figure 6. The structure of ResNet used in this paper

ResNet, proposed by He et al. [23], consists of a block with a "Shortcut Connection" and a down-sampling ResNet building block. A 1×1 convolution operation is added to the main branch of the downsampling building block. The ResNet 34-layer model is used. The network structure is the same as that of the original model except for the last fully connected layer with softmax. The number of nodes in the last fully connected layer is set to the number of classes. The pretrained model on ImageNet is used for transfer learning.

For classifier fusion, three models are performed by 300 epochs and using the batch size of 64; The learning rate is 0.001. To improve the performance of the three base architectures, data augmentation is introduced. For each original image, we generate a horizontally flipped image. We apply the following five image processing techniques: scaling, random translation, image rotation, and color brightness. In detail, we randomly scale the data with different sizes. Then, we randomly horizontally flip the samples. Additionally, we set up rotation to enhance the data. In addition, we generate different samples with different brightness, contrast and color brightness.

# 5. EXPERIMENTS

To validate the performance and robustness of the database, we implement PyTorch for CNN, LeNet, ResNet, AlexNet and decision-level classifier fusion. We divide the database into training and test sets. In the training set, 2,658 images are used, while for the testing set, 776 images used.

## 5.1 Baseline algorithms

We compare our proposed method with three classical algorithms: local binary pattern (**LBP**) [24, 25], local Gabor binary pattern (**LGBP**) [26], and convolutional neural network (**CNN**). The details are described as follows:

**LBP**: It is widely used in texture classification [24] and handwritten character classification [25], as LBP is a gray-scale invariant texture primitive statistic. The thresholded binary value of neighboring points against the center pixel is calculated, followed by multiplying with weights of the corresponding pixels and by summing up the result. Finally, the LBP code is obtained. To represent the feature of the character, we apply the LBP operator with $P$ equally spaced pixels on a circle of radius $R$, for each pixel. In the experiment, we consider $(1, 8)$, $(2, 16)$, and $(3, 24)$ for $(P, R)$. The uniform pattern is applied.

**LGBP:** This is a typical two-layer feature representation approach. The local Gabor magnitude as the first hidden layer was used for the purpose of enhancing image features, and then the LBP operator as the second latent layer was used to describe the image features by local structural information. In the experiment, the wavelength of Gabor is ranged from 4 to 8, and its orientation ranged from 0 to $\frac{\pi}{2}$. The same parameter of LBP is used on 32 Gabor magnitude maps for LGBP.

**CNN**: A CNN model is constructed for comparison. Here, we constructed three basic models, including 7-layer, 9-layer, and 11-layer architectures. The 7-layer model comprises of one input layer, one output layer, one fully connected layer, 2 convolutional layers, and 2 max-pooling layers. Based on the 7-layer model, 1 convolutional layer and 1 max-pooling layers is further added for the 9-layer model, while an additional 2 convolutional layers and 2 max-pooling layers is added for the 9-layer model. The filter size of the convolution layer is set as $3 \times 3$. The pooling size is set as 2. For simplification, we name the 7-layer, 9-layer, and 11-layer models as CNN-7, CNN-9, and CNN-11, respectively.

## 5.2 Algorithm comparison

We extract LBP and LGBP features with different parameters. We then classify the resulting feature vectors with a Support Vision Machine (**SVM**) based on a linear kernel [27]. The results are shown in Table 2. As shown in Table 2, with $R = 3$ and $P = 24$, LBP obtains recognition rate of 47.68%, while for LGBP the recognition rate of 42.34% with R= 2 and $P = 16$. Both methods fail to yield promising results. Instead, our proposed DCF-LAR obtains the recognition rate of 84.8%. DCF-LAR outperforms two handcrafted feature extractors.

Furthermore, to demonstrate the superiority of DCF-LAR, we compare it with three deep learning models based on convolutional neural network (CNN-7, CNN-9, and CNN-11) and three state-of-the-art deep architectures (LeNet, AlexNet, and ResNet), as the deep learning method can automatically extract the salient features that are invariant and a certain degree to shift and shape distortions of the input characters. Table 3 summarizes the obtained results. CNN-7, CNN-9, and CNN-11 obtain the recognition rates of 39.87%, 61.88%, and 55.23%, respectively. Compared with CNN-7, CNN-9, and CNN-11, our proposed DCF-LAR has significant increases in performance of 44.93%, 22.92%, and 29.57%, respectively.

Finally, among the baseline algorithms, the sole model including LeNet, AlexNet, and ResNet are trained by 4,000 epochs, 2,000 epochs, and 600 epochs, respectively. ResNet obtains a recognition rate of 82.17%**.** The recognition rate increases from 82.17% to 84.82% by using DCF-LAR. DCF-LAR outperforms ResNet. Table 2 clearly shows that DCF-LAR is more powerful than CNN, AlexNet, LeNet, and ResNet for extracting high-quality visual features from Houma Alliance Book ancient handwritten character images.

Table 2. Classification results in terms of recognition rate (%) for each traditional texture descriptor and ours. The best result is in bold.

| Methods | LBP | | | LGBP | | | Ours |
|---|---|---|---|---|---|---|---|
| | (1, 8) | (2, 16) | (3, 24) | (1, 8) | (2, 16) | (3, 24) | DCF-LAR |
| Accuracy | 37.50 | 45.49 | 47.68 | 35.58 | 42.34 | 39.48 | **84.82** |

Table 3. Performance comparison in terms of recognition rate (%) with deep neural networks. The best result is in bold.

| Method | CNN-7 | CNN-9 | CNN-11 | LeNet | AlexNet | ResNet | Ours |
|---|---|---|---|---|---|---|---|
| Accuracy | 39.87 | 61.88 | 55.23 | 62.72 | 78.87 | 82.17 | **84.82** |

## 5.3 Ablation study results

A comprehensive study is performed to evaluate our proposed method and understand the performance of our approach by removing each component of our method to fully comprehend the overall approach. Table 5 shows the ablation study results for our new database. We also show some of our results in Figure 7. As seen from Figure 7, for three examples, our proposed DCF-LAR accurately predicts the true result even in variants of the same class. We obtain the baseline results of using the sole model. The baseline results in terms of recognition rate are 62.72%, 78.87%, and 82.17% for DCF-L, DCF-A, and DCF-R, respectively.

Compared with DCF-L, fusing AlexNet, DCF-LA promisingly increases the performance by 17.7%, while fusing ResNet, DCF-LR significantly obtains better performance than DCF-LA. This demonstrates that fusing ResNet may provide a more accurate class probability for decision-level classifier fusion. Furthermore, it is interesting that DCF-AR obtains the highest recognition rate of 85.51%, which is even better than DCF-LAR. In contrast, DCF-LAR obtains the second-best recognition rate of 84.82%. Adding LeNet may not lead to increased performance. This may be explained by that the classifier probability generated by LeNet will cause confusion to the soft voting approach. However, DCF-LAR still competes with DCF-AR. Additionally, even compared with the sole deep neural network, our performance is qualitatively on par with the state of the art.

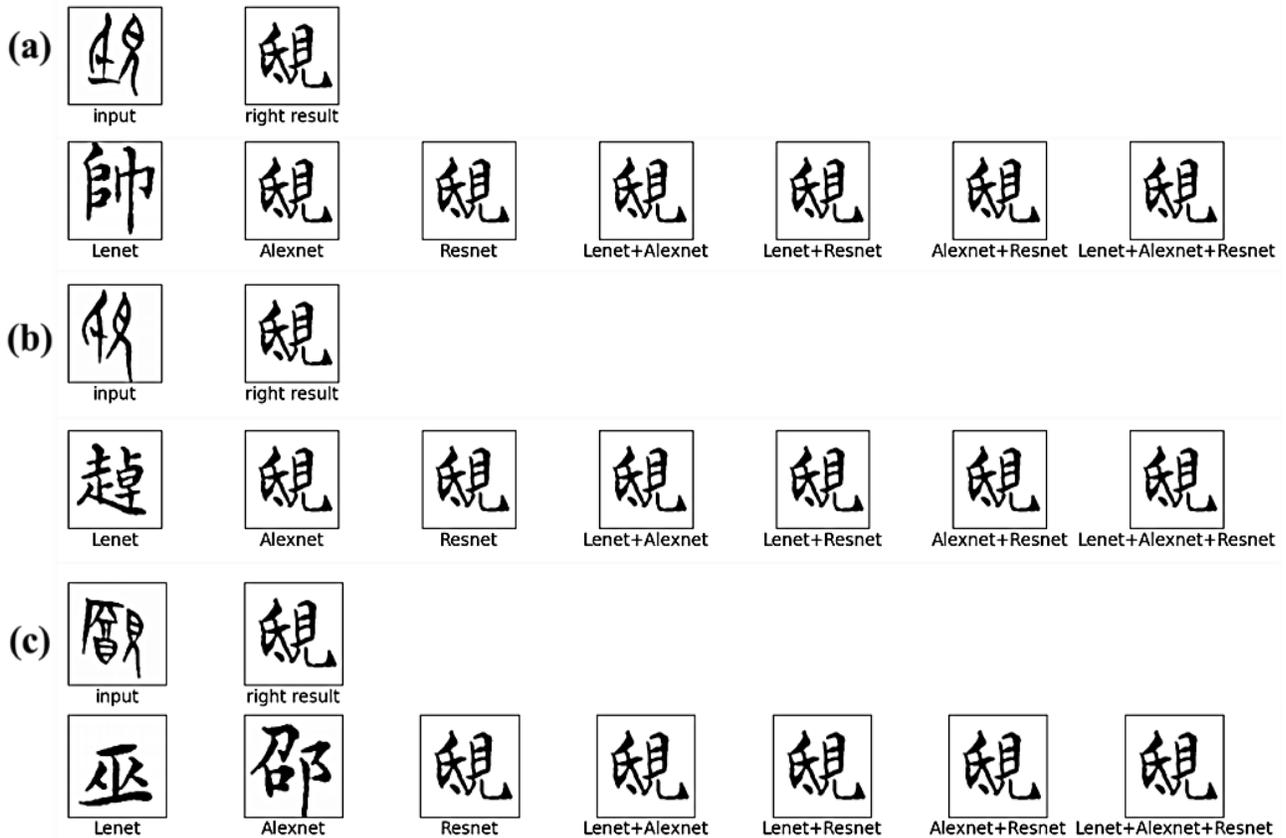

Figure 7. Example results on our new dataset.

Table 5. Decision-level classifiers fusion under combining different models in terms of recognition rate (%), where L, A, and R represent LeNet, AlexNet, and ResNet, respectively. DCF-AB represents DCF combining the models A and B. The best result is in bold.

| Method | LeNet | AlexNet | ResNet | Accuracy |
| --- | --- | --- | --- | --- |
| DCF-L | ✓ | | | 62.72 |
| DCF-A | | ✓ | | 78.87 |
| DCF-R | | | ✓ | 82.17 |
| DCF-LA | ✓ | ✓ | | 80.42 |
| DCF-LR | ✓ | | ✓ | 84.73 |
| DCF-AR | | ✓ | ✓ | **85.51** |
| DCF-LAR | ✓ | ✓ | ✓ | 84.82 |

## 6. CONCLUSION

In this work, we presented a new database of Houma Alliance Book ancient handwritten characters. The database contains 297 classes and 3,547 images from the archaeological report on the Houma Alliance Book. The ground truth was provided based on two forms of annotation, including page index and simplified Chinese character. Additionally, we presented a new method based on decision-level classifier fusion for the new database, which provides baseline results for Houma Alliance Book ancient handwritten character recognition.

In the future work, as the database suffers from the class imbalance problem, we will address that issue by proposing the semi supervised model. Moreover, we will investigate the stroke based recognition algorithm for Houma Alliance book ancient character recognition. Additionally, it is hoped that this database will be helpful to the research of digitized Houma Alliance book.

## ACKNOWLEDGEMENT

This research was in part supported by the National Natural Science Foundation of China (Grant NO. 62076122, 72001040), the Jiangsu Specially-Appointed Professor Program, the Talent Startup project of NJIT (NO. YKJ201982), the Opening Project of Jiangsu Province Engineering Research Center of IntelliSense Technology and System (NO. ITS202102), and the Opening Project of Advanced Industrial Technology Research Institute, Nanjing Institute of Technology (NO. XJY202102).

## REFERENCES

[1] C. Liu, F. Yin, D. Wang, and Q. Wan. CASIA Online and Offline Chinese Handwriting Databases. International Conference on Document Analysis and Recognition, pp. 37-41, 2011.

[2] Z. Wang, Y. Yu, Y. Wang, H. Long, and F. Wang. Robust end-to-end offline Chinese handwriting text page spotter with text kernel. International Conference on Document Analysis and Recognition, pp. 21-35, 2021.

[3] W. Wang, J. Zhang, J. Du, Z. Wang, Y. Zhu. DenseRAN for offline handwritten Chinese character recognition. International Conference on Frontiers in Handwritting Recognition, pp.104-109, 2018.

[4] Y. Chherawala, H. Dolfing, R. Dixon, and J. Bellegarda. Embedded large-scale handwritten Chinese character recognition. International Conference on Acoustics, Speech and Signal Processing, pp. 8169-8173, 2020

[5] M. Guo. A Probe into Houma's League Book. A Collection of Research Papers on Houma's League Books. Shanxi Publishing House Media Group Sanjin Publishing House, pp.8-13, 2015.

[6] R. Casey and G. Nagy. Recognition of Printed Chinese Characters. IEEE Transactions on Electronic Computers, vol. 15, no. 1, pp. 91-101, 1966.

[7] D. Ciresan and U. Meier. Multi-column deep neural networks for offline handwritten Chinese character classification. International Joint Conference on Neural Networks, pp. 1-6, 2013.

[8] X. Liu, B. Hu, Q. Chen, X. Wu, and J. You. Stroke sequence-dependent deep convolutional neural network for online handwritten Chinese character recognition. IEEE Transactions on Neural Networks and Learning Systems, vol. 31, no. 11, pp. 4637-4648, 2020.

[9] L. Lin, X. Wang, and B. Liu. Combining multiple classifiers based on statistical method for handwritten Chinese character recognition. International Conference on Machine Learning and Cybernetics, pp. 252-255, 2002.

[10] H. Wang and H. Shi. A textual research on Houma alliance. Cultural relics world, vol. 01, pp. 83-98, 1996.

[11] D. Zhang. The Houma Alliance Book - List of Characters (Revised). Journal of Xinyi National Teacher's College, vol. 01, pp. 40-45, 2014.

[12] S. Lin, S. Zhang, and X. Shi. Virtual restoration method of Houma convenant tablets based on suprathreshold stochastic resonance. Journal of Graphics, vol. 38, no. 3, pp. 361-366, 2022.

[13] A. Zoizou, A. Zarghili, and I. Chaker. MOJ-DB: a new database of Arabic historical handwriting and a novel approach for subwords extraction. Pattern Recognition Letters, vol. 159, pp. 54-60, 2022.

[14] B. Li, Q. Dai, F. Gao, W. Zhu, Q. Li and Y. Liu. HWOBC- A handwriting oracle bone character recognition database. Journal of Physics: Conference Series, vol. 1651, 012050, 2020.

[15] S. Narang, M. Jindal, S. Ahuja, and M. Kumar. On the recognition of Devanagari ancient handwritten characters using SIFT and Gabor features. Soft Computing, vol. 24, pp. 17279-27289, 2020.

[16] H. Zhao, H. Chu, Y. Zhang, and Y. Jia. Improvement of ancient Shui character recognition based on convolutional neural network. IEEE Access, vol. 8, pp. 33080-33087, 2020.

[17] Y. Xu, F. Yin, D. Wang, X. Zhang, Z. Zhang, and C. Liu. CASIA-AHCDB: a large-scale Chinese ancient handwritten characters database. International Conference on Document Analysis and Recognition, pp. 793-798, 2019.

[18] F. Cheng. The Houma and Wenxian alliance books[J]. Journal of Yindu Studies, 2002,(04):46-49.

[19] X. Huang, G. Zhao, M. Pietikainen, and W. Zheng. Dynamic facial expression recognition using boosted component-based spatiotemporal features and multi-classifier fusion. Advanced Concepts for Intelligent Vision System, pp. 312-322, 2010.

[20] Y. Huang and C. Suen. A method of combing multiple experts for the recognition of unconstrained handwritten numerals. IEEE Transactions on Pattern Analysis and Machine Intelligence, vol. 17, no. 1, pp. 90-94, 1995.

[21] Y. Cun, B. BOser, J. Denker, D. Henderson, R. Howard, W. Hubbard, and L. Jackel. Handwritten digit recognition with a back-propagation network. Neural Information Processing System, vol. 2, pp. 396-404, 1990.

[22] A. Krizhevsky, I. Sutskever, and G. Hinton. ImageNet classification with deep convolutional neural networks. Advances in Neural Information Processing Systems, vol. 25, pp. 1-9, 2012.

[23] K. He, X. Zhang, S. Ren, and Jian Sun. Deep Residual Learning for Image Recognition. IEEE Conference on Computer Vision and Pattern Recognition, pp. 770-778, 2016.

[24] T. Ojala, M. Pietikainen, and T. Maenpaa. Multiresolution gray-scale and rotation invariant texture classification with local binary patterns. IEEE Transactions on Pattern Analysis and Machine Intelligence, vol. 24, no. 7, pp. 971-987, 2002.

[25] F. Anggraeny, E. Mandyartha, and D. Kartika. Texture feature local binary pattern for handwritten character recognition. International Conference on Information Technology International Seminar, 2020.

[26] W. Zhang, S. Shan, W. Gao, X. Chen, and H. Zhang. Local Gabor binary pattern histogram sequence (LGBPHS): a novel non-statistical model for face representation and recognition. IEEE International Conference on Computer Vision, pp. 1-5, 2005.

[27] C. Chang and C. Lin. LIBSVM: a library for support vector machines. ACM Transactions on Intelligent Systems and Technology, vol. 2, no. 27, pp. 1-27, 2011.